\documentclass[10pt,final,twocolumn]{IEEEtran}

\usepackage{textcomp}
\DeclareSymbolFont{tildelow}{TS1}{cmr}{m}{n}
\DeclareMathSymbol{\tildelow}{0}{tildelow}{126}

\usepackage{bbm}
\usepackage{xcolor}
\usepackage{color}
\usepackage{graphicx}
\usepackage{epsfig}
\usepackage{subfigure}
\usepackage{amssymb}
\usepackage{amsbsy}
\usepackage{cite}
\usepackage{amsthm}
\usepackage{romannum}
\usepackage{algorithm}

\usepackage[margin=0.95in]{geometry}

\usepackage{float}
\usepackage[T1]{fontenc}
\usepackage{url}
\usepackage{ifthen}
\usepackage{cite}
\usepackage[cmex10]{amsmath} 
\usepackage{tikz}

\usepackage{physics}
\usepackage{algcompatible}

\newcommand{\K}{\mathbb{K}}
\newcommand{\s}{\mathcal{S}}
\newcommand{\al}{\mathcal{A}}
\newcommand{\gi}{\mathcal{G}}
\newcommand{\Hi}{\mathcal{H}}
\newcommand{\A}{\mathbb{A}}
\newcommand*\mean[1]{\overline{#1}}
\usepackage{mathtools}
\DeclarePairedDelimiter\ceil{\lceil}{\rceil}

\newtheorem{theorem}{Theorem}

\usepackage{dsfont}
\DeclareMathOperator*{\argmax}{arg\,max}
\DeclareMathOperator*{\argmin}{arg\,min}
\usepackage{enumitem}
\usepackage{float}
\begin{document}

\title{Client Selection for Generalization in Accelerated Federated Learning: \\ A Multi-Armed Bandit Approach}

\author{
  {Dan Ben Ami, Kobi Cohen, Qing Zhao}
	\thanks{
		D. Ben Ami and K. Cohen are with the School of Electrical and Computer Engineering, Ben-Gurion University of the Negev, Beer-Sheva, Israel (e-mail:danbenam@post.bgu.ac.il; yakovsec@bgu.ac.il).} 
		\thanks{
	Qing Zhao is with the Department of Electrical and Computer Engineering, Cornell University, Ithaca, New York, USA (e-mail: qz16@cornell.edu).
	}
\thanks{A short version of this paper was accepted for presentation in the IEEE International Conference on Acoustics, Speech and Signal Processing (ICASSP) 2023 \cite{benami2023client}.}
\thanks{This research was supported by the ISRAEL SCIENCE FOUNDATION (grant No. 2640/20)}
	\vspace{-0.75cm}
}

\maketitle
\pagenumbering{arabic}
\begin{abstract}
\label{sec:abstract}
Federated learning (FL) is an emerging machine learning (ML) paradigm used to train models across multiple nodes (i.e., clients) holding local data sets, without explicitly exchanging the data. It has attracted a growing interest in recent years due to its advantages in terms of privacy considerations, and communication resources. In FL, selected clients train their local models and send a function of the models to the server, which consumes a random processing and transmission time. The server updates the global model and broadcasts it back to the clients. The client selection problem in FL is to schedule a subset of the clients for training and transmission at each given time so as to optimize the learning performance. In this paper, we present a novel multi-armed bandit (MAB)-based approach for client selection to minimize the training latency without harming the ability of the model to generalize, that is, to provide reliable predictions for new observations. We develop a novel algorithm to achieve this goal, dubbed Bandit Scheduling for FL (BSFL). We analyze BSFL theoretically, and show that it achieves a logarithmic regret, defined as the loss of BSFL as compared to a genie that has complete knowledge about the latency means of all clients. Furthermore, simulation results using synthetic and real datasets demonstrate that BSFL is superior to existing methods.
\end{abstract}

\begin{IEEEkeywords}
Federated learning (FL), client selection, client scheduling, multi-armed bandit (MAB), generalization in machine learning. \vspace*{-0.2cm}
\end{IEEEkeywords}
\section{Introduction}
\label{sec:introduction}
The increasing demand for ML tasks in wireless networks consisting of a large number of clients (i.e., users or edge devices) has led to the rise of a new ML framework, called federated learning (FL) \cite{mcmahan2017communication, bonawitz2019towards, gafni2022federated}. FL enables to train ML models without extracting the data directly from the edge devices. In the beginning of the FL procedure, the ML model is being initialized. Next, the FL training scheme is done by repeated iterations between the clients that implement local training and the parameter server that aggregates functions of the local trained models. This allows to train the global model without explicitly exchanging the data, which is advantageous in terms of privacy considerations and communication resources. The bandwith constraint dictates the total number of clients that can be scheduled for transmissions, which can be implemented by traditional digital communications over orthogonal channels \cite{srikant2013communication, bonawitz2019towards, gafni2022federated} or coherent analog transmissions over multiple access channels \cite{amiri2020federated, amiri2020machine, sery2019sequential, sery2020analog, sery2021over, paul2021accelerated}. To simplify the presentation, we focus here on traditional digital communications.

\begin{figure}[h!]
\centering \epsfig{file=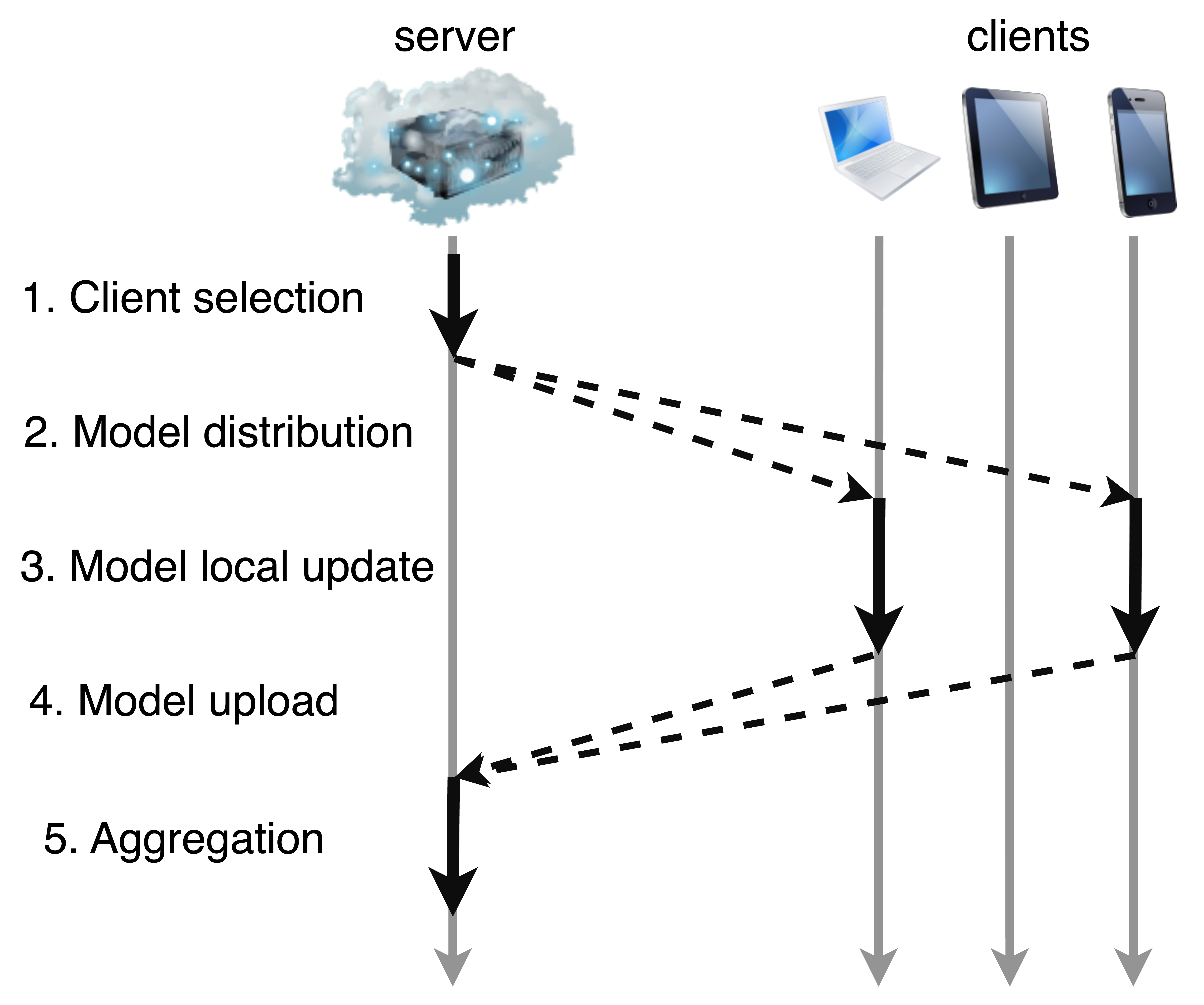,
width=0.3\textwidth}
\vspace{-0.3cm} 
\caption{An illustration of the FL communications scheme.} \label{fig:FL iter}
\end{figure}

We consider an FL system with $K$ clients sharing $m$ orthogonal channels at each iteration (e.g., OFDMA), where $m\leq K$. Since the bandwidth is limited by $m$ channels and the number of clients $K$ is typically high, only a small fraction of the clients can be scheduled for transmission at each iteration \cite{konevcny2016federated}. The selection of a subset of $m$ clients among $K$ clients has been studied in the context of scheduling and spectrum access problems in traditional communication schemes (see e.g., \cite{tekin2012online, liu2012learning, srikant2013communication, naparstek2018deep, bistritz2018distributed, gafni2020learning,  gafni2022distributed, gafni2021learning}). However, solving the problem in FL systems adds new challenges resulting in fundamentally different algorithms and analysis \cite{gafni2022federated}. Each iteration in the FL system includes selecting a subset of $m$ clients, distributing the model from the server to the selected clients, training locally the model by the selected clients, uploading the trained models from the clients to the server (i.e., sharing $m$ orthogonal uplink channels using OFDMA), and finally aggregating the received models at the server to update the global model. An illustration is presented in Fig. \ref{fig:FL iter}.
New emerging challenges in the FL framework are described next \cite{kairouz2021advances, li2020federated, gafni2022federated}. The first challenge is the heterogeneous data between the clients \cite{li2020federated, sery2021over}. The local data is usually subjective to the client, and therefore is likely to be biased and imbalanced. Another challenge is balancing the total latency affected by different clients with different computational and communication resources, as well as different channel states. This is a key challenge in FL systems since each iteration ends when the server receives the trained models from all scheduled clients, each experiences a different random latency. In this paper, we tackle these challenges and develop a novel client selection algorithm which is superior to existing methods, and achieves strong performance in terms of reducing the training latency, while not harming the generalization of the model.

\subsection{Client Selection Methods in Federated Learning}
\label{ssec:client}

Previous studies have shown that client selection has a great impact on the model's performance and the training latency. The standard client selection method is to choose the clients randomly with probability proportional to the amount of data each client contains \cite{mcmahan2017communication, li2019fair}. The clients can differ from each other in several factors such as the size of their database, the amount of available resources, their channel state, the energy they can invest, and the value of their local loss function. Each iteration in FL systems ends when the last scheduled client uploads its updated model. Thus, variability between clients might lead to poor performance of the standard random selection method. As a result, more recent methods used observations of one or more of the aforementioned factors to infer the iteration latency of each client and subsequently select the
appropriate group that leads to efficient training latency \cite{nishio2019client, xu2020client, abdulrahman2020fedmccs, rizk2022federated, yang2020age, ozfatura2021fast}. In \cite{nishio2019client, xu2020client, abdulrahman2020fedmccs, rizk2022federated} the focus was on the effect of client sampling on the learning performance. In \cite{yang2020age, ozfatura2021fast} the problem is investigated under the assumption that parameters related to the communication link and latency are known (e.g., channel state, transmission latency). As a result, the optimization problem in those studies is deterministic with respect to the client selection strategy. However, in many scenarios, client latencies are affected by other random factors as discussed earlier (e.g., client energy state, available computational resources, random transmission rate), and are unknown at the time of the client selection. Therefore, recent studies (including this paper) tackle the problem using a fundamentally different approach by treating clients' latencies as unknown random variables drawn from unknown distributions, which leads to a stochastic optimization problem that raises the well-known exploration versus exploitation dilemma. On one hand, it is necessary to explore different actions (i.e., client selection sets) to explore the system state. On the other hand, it is imperative to exploit the information gathered so far to converge to the optimal selection strategy. This approach involves modeling and analyzing the problem as a MAB problem by treating the iteration latency or the local loss function of each client as a reward taken from an unknown distribution and given to the server (modeled as the gambler in MAB) \cite{xia2020multi, yoshida2020mab, cho2020bandit, xu2021online, huang2020efficiency}. In \cite{xia2020multi}, the authors developed a naive MAB method as in the classic i.i.d. MAB that converges to a strategy that selects a small subset of the quickest clients (i.e., with the smallest expected latency). This method, however, suffers from over-fitting to the quickest clients' data when applying to scheduling in FL systems. In \cite{yoshida2020mab, cho2020bandit, xu2021online}, the focus was on reducing the loss, but the generalization issue remained open. Therefore, in \cite{huang2020efficiency} the authors suggested to use a fairness constraint to ensure that each client is selected in a certain proportion of the communication rounds. However, this method still converges to the group of the quickest clients except for the times of selecting the slower clients due to the fairness constraint. Therefore, we suggest a novel MAB formulation for the client selection problem consisting of a reward function that captures the generalization of the ML task. This allows us to solve the problem rigorously in this paper.

\subsection{Main Contributions}

To solve the client selection problem in FL, it is needed to solve an online learning problem with the well-known exploration versus exploitation dilemma. On the one
hand, the player (i.e., the server) should explore all arms (i.e., clients) in order to learn their state (i.e., latency distribution, generalization ability) which affects the overall training time of the FL task. On the other hand, it should exploit the information
gathered so far to select the most rewarding subset of arms at each given time. In this paper we develop a rigorous MAB formulation to model this problem, and a novel learning algorithm to solve it. We provide rigorous theoretical analysis of the algorithm, as well as extensive numerical analysis. Below, we summarize the main contributions.

\begin{itemize}
     \item \textbf{
A novel MAB formulation for the client selection
problem:} We present a new MAB formulation that trades-off between the training latency and the generalization of the model in FL by client selection. This trade-off raises new challenges in the learning design. On the one hand, it is desired to reduce the training process time of the model, and on the other hand, it is desired to increase the ability of the model to generalize by avoiding over-fitting. The novel MAB formulation tackles this trade-off by selecting clients based on a time-varying reward influenced by the history of previous selections.
    \item \textbf{Algorithm development:} We develop a novel upper confidence bound (UCB)-based algorithm to solve the problem, dubbed Bandit Scheduling for FL (BSFL). BSFL presents a fundamentally different approach of selecting the clients, as the reward function updates the contributions of each client to the FL task based on the history of previous selections. As a result, BSFL does not aim to converge to selecting a fixed subset of clients, but rather aims to increase the ability of the model to generalize. Furthermore, we provide concrete examples of the use of BSFL in both cases where the data is i.i.d. and balanced across clients, and where the data is non-i.i.d. and imbalanced. Since the UCB optimization in BSFL requires to solve a combinatorial problem to select the most rewarding subset of clients at each given time, the time complexity increases exponentially with the number of clients, which might be infeasible for a large number of clients. Hence, we utilize the optimization problem's structure to devise a low-complexity algorithm, named accelerated lightweight simulated annealing (ALSA), which relies on a newly proposed accelerated lightweight simulated annealing technique that we develop for BSFL. Specifically, we show analytically that ALSA reduces the degree-order in the simulated annealing graph from $O(K^2)$ to $O(K)$ (where $K$ is the number of clients), and still keeps the convergence property. This results in a significant accelerated convergence time requires to solve the UCB optimization in BSFL.
    \item \textbf{Performance analysis:} 
    We analyze the performance of BSFL theoretically and numerically. In terms of theoretical analysis, as commonly done in the MAB literature, we evaluate the performance by regret, defined as the loss of BSFL as compared to a genie that has complete knowledge about the latency means of all clients. We analyze the performance of BSFL rigorously, and show that it achieves a logarithmic regret with time. In terms of numerical analysis, we present extensive simulations using synthetic and real datasets. All simulations demonstrate that BSFL is superior to existing methods in both the regret and the prediction accuracy. 
\end{itemize}

\section{System Model}
\label{sec:system}

We consider a common FL system with star topology. The FL system consists of a set $\K$ of clients with cardinality $|\K|=K$, where the clients communicate directly with the server via $m$ orthogonal channels (e.g., OFDMA), where $m\leq K$. Since the bandwidth is limited by $m$ channels and the number of clients $K$ is typically high, only a small fraction of the clients can be scheduled for transmission at each iteration \cite{konevcny2016federated}. Due to operation or communication constraints, the server interacts with a set $\A_t \subseteq \K$ of clients which are available to participate in the FL task at iteration $t$. The number of available clients $|\A_t|$ is assumed to be much higher than the number of channels $m$. At each iteration, the server selects $m$ clients from the set $\A_t$ of clients to participate in the FL task, each client is assigned to a dedicated orthogonal channel. We denote the set of all possible client selections by $H(\K)=\{\s\subset \K: |\s|=m\}$, and the set of all possible client selections at iteration $t$ by $H(\A_t)$, i.e., $H(\A_t) = \{\s \subset \A_t: |\s|=m\} \subseteq H(\K)$.

Each client $k\in\K$ holds a local database $X_k$. The local data is not being shared in FL systems due to privacy or communication constraints, and only the output model of a local training procedure is transmitted to the server. The FL task by the server is to minimize the global loss function denoted by: 
\begin{equation}
    \mathcal{L}(W,X)=\sum_{k\in\K}\frac{|X_k|}{|X|}\cdot \mathcal{L}(W,X_k)
\end{equation} 
with respect to the model's parameters denoted as $W$, where $X=\cup_{k\in\K}X_k$ and $\mathcal{L}(W,X_k)$ is the local loss of client $k$. The server's action at iteration $t$ is defined by the selection of the participating set of clients. We denote this action in the algorithm that we aim to develop by $\al_t\subseteq H(\A_t) $. We also denote the history of all previous client selections by the server before iteration $t$ by $\mathcal{H}_t=\{\al_1,\al_2,...,\al_{t-1}\}$. 

Each iteration consumes a random processing time depending on the participating clients due to distributing the model to the clients, local training and updating the model at each client, and transmitting each local model back to the server for global aggregation. We denote $\tau_{k,t}$ as the total random latency consumed by client $k$ at iteration $t$. At each iteration $t$, the FL process proceeds to the next iteration after receiving all the trained models from all the selected clients. Therefore, the total iteration time $\tau_t$ is dictated by the slowest client. We further assume a fixed latency bound, $\tau_{max}$, where all clients must meet so that the server receives their local model. As a result, the total latency at iteration $t$ is given by:
\begin{equation}\label{eq: tau_t}
    \tau_t = \min\{\max_{k\in \al_t}{\tau_{k,t}}, \tau_{max}\} .
\end{equation}

\section{MAB-Based formulation for the client selection problem}
\label{sec: formulation}

In this section, we present a novel MAB formulation for the client selection problem in order to minimize the training latency without
harming the ability of the model to generalize, i.e., to provide reliable predictions for new observations. 

\subsection{Trading-off Between the Generalization and Iteration Time via MAB Formulation}

MAB problems are often illustrated with the example of a player or gambler facing a row of slot machines, selecting arms to pull at each time and obtaining rewards accordingly. In our context, the server acts as the player, and the clients act as the arms. At each iteration, the server selects a subset of clients, which contribute to the learning rate accordingly. Solving the MAB problem, and specifically the client selection problem in this paper, requires addressing the well-known exploration versus exploitation dilemma in online learning. On one hand, the server should explore all clients to learn their state, such as their latency distribution and generalization ability, which affect the overall training time of the FL task. On the other hand, the server should exploit the information gathered so far to select the most rewarding subset of clients at each given time.

The exploration versus exploitation dilemma has been rigorously addressed in MAB optimization \cite{zhao2019multi}. As a result, recent studies have modeled and analyzed the client selection problem in FL as a MAB problem, treating the iteration latency or local loss function of each client as a reward obtained from an unknown distribution and given to the server \cite{xia2020multi, yoshida2020mab, cho2020bandit, xu2021online, huang2020efficiency}, as discussed in Section \ref{ssec:client}. However, all these methods converge to selecting the fastest clients (except for the times of selecting slower clients due to the fairness constraint in \cite{huang2020efficiency}). Consequently, they suffer from overfitting to the data of the quickest clients when applied to scheduling in FL systems, resulting in decreased ability to generalize and decreased prediction accuracy.

In this paper, we overcome this limitation by developing a novel MAB formulation for the client selection problem consisting of a time-varying reward function that captures both the client latency as well as the client ability to contribute to the generalization of the ML task. It should be noted that previous work on rotting bandits \cite{levine2017rotting} considered a reward function that decreases with the activation rate of selected arms. However, the model here is fundamentally different, since the time-varying reward function depends on the overall subset selections of correlated clients in the systems. This leads to a novel combinatorial MAB (CMAB) formulation \cite{chen2013combinatorial}, in which the design and analysis is fundamentally different from rotting bandits. Specifically, our formulation trades-off rigorously between the generalization and latency, as each client in FL systems might have different characteristics such as computing power, task operations, channel states, etc. This results in different latency distributions in training and transmitting the local models across clients in the FL system \cite{gafni2022federated}. On the one hand, to reduce the total iteration time at each iteration, it is desired to select clients with small latencies at each iteration. On the other hand, it is desired to select as many clients as possible during the training process (even slower clients) for the model to be robust and generalized for both i.i.d. balanced data and non-i.i.d. imbalanced data \cite{hardt2016equality, mohri2019agnostic, huang2020efficiency, shi2021survey}. In the next susbection we describe the generalization function used in the CMAB formulation.

\subsection{The Generalization Function}

For simplicity and maintaining generality, we denote by $\Theta$ all system constant parameters that contribute to the reward (which will be described later
with examples). The level of contribution of client $k$ to the generalization ability and robustness of the model is evaluated by a generalization function $g_{k,\Theta}:\Hi_t\to[-1,1]$. It gives at each iteration $t$ a value that represents the contribution of client $k$ to the generalization of the global model, depending on the model constant parameters $\Theta$ and the selection history before time $t$, $\Hi_t$. We next present concrete examples of the function $g_{k,\Theta}$ in both cases of i.i.d. and balanced data, and non-i.i.d. and imbalanced data across clients. 

We start by considering the case of i.i.d. balanced data. We use the selection history to extract a counter $c_{k,t}$ for each client that counts the number of iterations that client $k$ was selected until the beginning of iteration $t$, i.e., $c_{k,t} = \sum_{i=1}^{t-1}\mathbbm{1}_{\{k\in \al_i\}}$. In order to achieve high generalization in this case, it is important that each client is selected an equal number of times \cite{hardt2016equality, mohri2019agnostic, li2019fair, huang2020efficiency, shi2021survey}. In this case, we let $\Theta$ store the number of clients and the number of channels, i.e., $\Theta=\{K,m\}$. Then, an effective design of the function $g_{k,\Theta}$ is given by:
\begin{equation}
    g_{k,\Theta}(\Hi_t) = g_{k,\Theta}\Bigl(\frac{c_{k,t}}{t}\Bigr) = \biggl|\frac{m}{K}-\frac{c_{k,t}}{t}\biggr|^\beta \cdot sgn\biggl(\frac{m}{K}-\frac{c_{k,t}}{t}\biggl),
\end{equation}
where $\beta$ is a tuning parameter (natural number), and $sgn(\cdot)$ is the sign function that returns $\pm1$. An illustration is presented in Fig. \ref{fig:g_drew}. 

The key idea behind the design is that the more the ML model is trained with new data points or alternatively with data that was used too little, the more its ability to generalize increases \cite{hardt2016equality, shi2021survey}. It is thus desired to select clients uniformly with rate $m/K$ over time. Therefore, the generalization function $g_{k,\Theta}$ is designed to be monotonically decreasing so that it provides incentive (i.e., positive value in terms of contributing to generalization) to select clients that were selected too little (i.e., $\frac{c_{k,t}}{t}< m/K$), and returns zero when the client's counter reaches the desired selection rate (i.e., $\frac{c_{k,t}}{t}=m/K$). Clients that were selected too frequently are given negative values in terms of contributing to generalization. 

\begin{figure}[h!]
\centering \epsfig{file=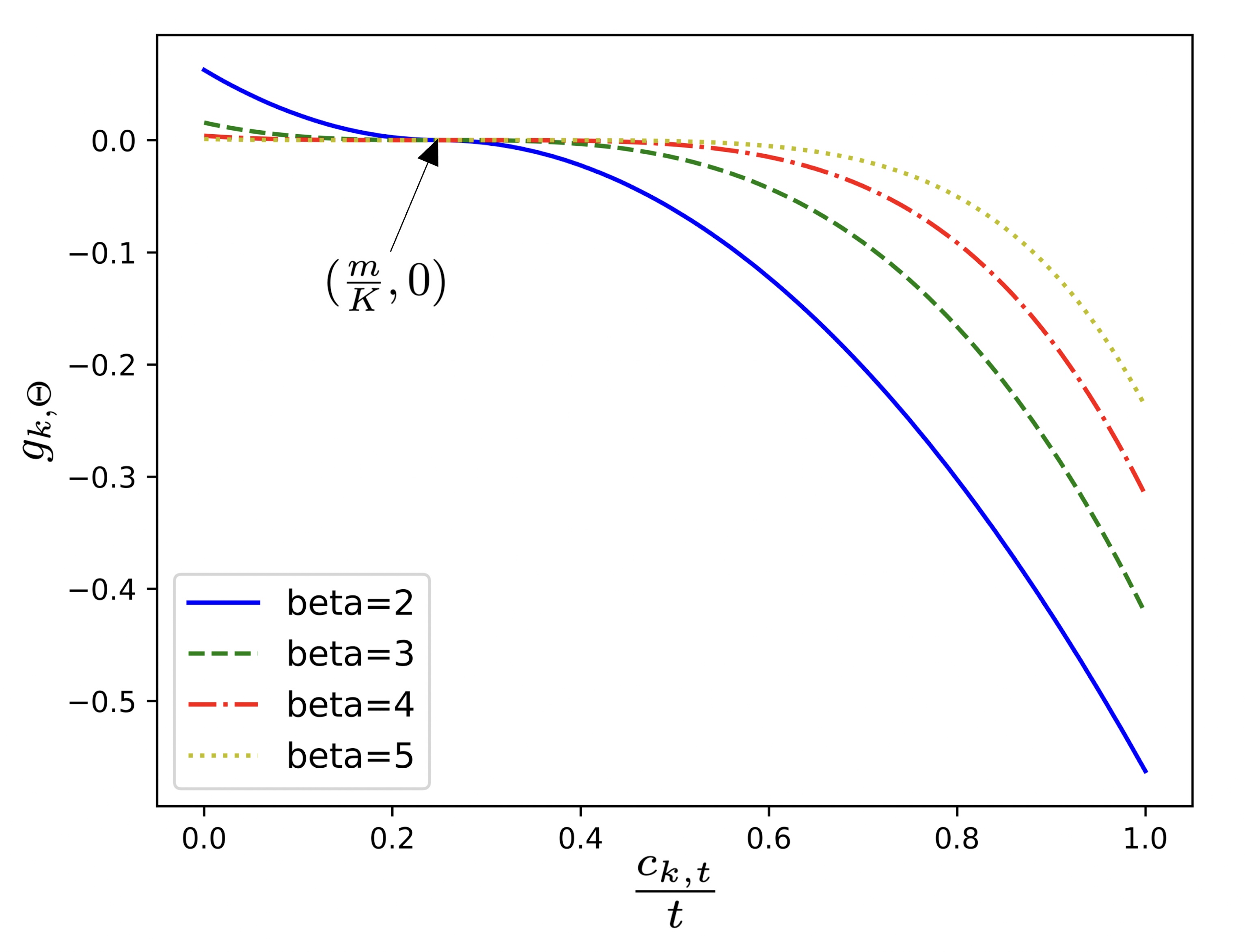,
width=0.45\textwidth}
\vspace{-0.2cm} 
\caption{The value of $g_{k,\Theta}$ as a function of $\frac{c_{k,t}}{t}$ with different values of $\beta$.} 
\label{fig:g drew}
\end{figure}

In the second scenario, consider non-i.i.d. and imbalanced data, where the data quality varies between clients. In this case, it is desired to take into account the participation rate of each client, as well as the quantity and quality of the data, when selecting clients for transmission. The quantity of each client's data (i.e., its database size $|X_k|$) is useful for the generalization, as it more likely to represent the true distribution of data \cite{xu2012robustness}. The second aspect stems from the desire to train the model on reliable, high-quality, and meaningful data \cite{zhu2004class, atla2011sensitivity, nettleton2010study, gupta2019dealing}.
Generally, in ML, and particularly FL, the data for each client might have different quality due to several factors such as device heterogeneity, environment heterogeneity, etc. We define the client's data quality by $q_k\in[0,1]$, where $1$ stands for the best quality, and thus the generalization function $g_{k,\Theta}$ is designed to prioritize clients with higher data quality. Since the quality for each client (say client $k$) is aggregated over all its data points $X_k$, we denote the overall significance of client $k$ by $d_k = q_k \cdot |X_k|$. Therefore. the model parameters in this case is given by $\Theta = \bigl\{K,m,\{(|X_k|, q_k): k\in\K\}\bigl\}$, and the generalization function is given by:
\begin{equation}
    g_{k,\Theta}(c_{k,t}) = \biggl|\frac{m d_k}{\sum_{k\in\K}{d_k}}-\frac{c_{k,t}}{t}\biggr|^\beta \cdot sgn\biggl(\frac{m d_k}{\sum_{k\in\K}{d_k}}-\frac{c_{k,t}}{t}\biggl) .
\end{equation}

In the next subsection we present the CMAB formulation for the client selection problem. In the empirical study in this paper, we used the forms of $g_{k,\Theta}$ as detailed above, which demonstrated very good performance. It should be noted, however, that the CMAB formulation is general. It does not depend on a specific form of $g_{k,\Theta}$, and other forms can be used. For example, it is possible to define $g_{k,\Theta}$ to be dependent on the local loss value of each client, which is advantageous in improving the convergence in the long term \cite{cho2022towards}. 

\subsection{CMAB-Based Formulation with History-Dependent Reward}

We now formulate the client selection problem as a CMAB problem with history-dependent reward. At each iteration $t$, the server (i.e., the player) selects a subset of clients $\al_t \in H(\A_t)$ (i.e., arms), and at the end of the iteration observes the clients' iteration latencies. The reward given at the end of iteration $t$ is defined by:
\begin{multline}
\label{eq:reward}
        r(t) = \frac{\frac{\tau_{min}}{\tau_{max}}}{\max_{k\in \al_t}{\frac{\tau_{k,t}}{\tau_{max}}}}+\alpha\cdot\frac{1}{m}\sum_{k\in \al_t}g_{k,\Theta}(\Hi_t) \\
        = \min_{k\in \al_t}{\{\frac{\tau_{min}}{\tau_{k,t}}\}}+\frac{\alpha}{m}\sum_{k\in \al_t}g_{k,\Theta}(\Hi_t),
\end{multline}
where $\alpha$ is a hyper-parameter of the generalization, balancing between the amount of the generalization and the iteration time, $\tau_{min}$ is the minimum iteration latency possible, which is the time needed to send and receive the model, as if the client required zero training time, and $\tau_{max}$ is the fixed latency bound as defined before (\ref{eq: tau_t}). For purposes of analysis, we further assume that the reward given at the end of each iteration is quantized and the smallest difference between two different rewards is denoted by $\Delta_{min}$, as commonly assumed in analyzing MAB-based problems. Since the iteration time is determined by the maximal latency under the selected subset of clients $\al_t$, the first term on the RHS of \eqref{eq:reward} represents the reward due to the iteration time, and the second term represents the reward due to the generalization. The smaller the value of $\alpha$, the higher priority to select faster clients. On the one hand, 
this reduces the iteration time. On the other hand, it tends to select the same faster clients at each iteration, which reduces the generalization, and increases the total number of iterations needed to achieve reliable predictions.

An important insight from the CMAB formulation is that if a slow client is given high priority based on the reward function and is selected for transmission at the current iteration, then since the iteration time is determined by the slowest client, there is no motivation to choose fast clients at the same iteration. Instead, clients with high rewards in terms of generalization, even if they are slow, should be prioritized.

Note that the formulation of our CMAB problem is fundamentally different from the classical CMAB problem. Specifically, here the reward does not depend on the new observations solely (as in classic CMAB), but also on the selection history up to the current iteration. Thus, optimizing the reward in classic MAB leads to a fixed selection of arms, while in our problem the optimal selection is time-varying depending on the selection history, as will be detailed in the next subsection.

\subsection{The Objective}
\label{objective}

We aim to find the best selection policy $\Pi=\{\al_1,\al_2,...\}$ (where $\al_t$ stands for the selection at iteration $t$) which minimizes the training process time while preserving the ability of the model to perform the generalization. The performance of online learning algorithms are commonly evaluated by the regret, defined as the loss of an algorithm as compared to genie with side information on the system. Here, to measure the performance of our proposed policy we define the regret as follows. We denote the maximal mean reward at iteration $t$ that can be obtained by genie that has complete knowledge about the latency 
means of all clients by $r^*(t)$, where genie's selection at time $t$ is given by:
\begin{equation}
\label{Gt}
      \gi_t = \argmax_{\s \in H(\A)}\biggl\{\min_{k\in \s}{\mu_k}+\frac{\alpha}{m}\sum_{k\in \s}g_{k,\Theta}\bigl(\Hi_t\bigl)\biggl\},
\end{equation}
where $\mu_k$ stands for the speed mean of client $k$ (i.e., $\mu_k = \mathbb{E}\bigl[\frac{\tau_{min}}{\tau_k}\bigl]$). 

The regret of the policy (say $\Pi$) at time $n$ is defined by the accumulated loss by time $n$ between the reward obtained by genie and the expected reward obtained by $\Pi$:

\begin{equation}
\label{eq:regret}
    R(n) = \mathbb{E}_\Pi\biggr[\sum_{t=1}^n{ r^*(t)-r(t)}\biggr].
\end{equation}

The objective is thus to find a policy $\Pi$ that minimizes the regret order with time.

\section{The Proposed BSFL Algorithm}
\label{algo}

In this section we start by presenting the BSFL algorithm to solve the objective. Then, we analyze the regret (\ref{eq:regret}) analytically. 

\subsection{Description of the Algorithm}

The BSFL algorithm is based on a novel UCB-type design for client selection in FL. The pseudocode for the BSFL algorithm is provided in Algorithm \ref{alg:Algorithm}. We now discuss the steps of the BSFL algorithm in detail. 

BSFL observes realizations of the random speed of the selected clients (i.e., $\frac{\tau_{min}}{\tau_{k,t}}$) and estimates its mean accordingly. Let
\begin{equation}
    \mean{\mu_{k,t}} = \frac{1}{c_{k,t}}\sum_{i=1}^{t-1}\frac{\tau_{min}}{\tau_{k,i}}\cdot\mathbbm{1}_{\{k\in \al_i\}}
\end{equation} 
be the sample-mean speed of client $k$ after $t$ iterations. We design the UCB function of client $k$'s speed after $t$ iterations by:
\begin{equation}
    \mbox{ucb}(k,t) = \mean{\mu_{k,t}}+\sqrt{\frac{(m+1)\ln{t}}{c_{k,t}}}.
\end{equation}

To maintain an updated $\mean{\mu_{k,t}}$ and $\mbox{ucb}(k,t)$ for each client, each time a client is selected to participate in the FL iteration, the algorithm observes its speed $\frac{\tau_{min}}{\tau_{k,t}}$ and updates its counter and the speed's sample-mean as follows:
\begin{equation}
\label{update_c}
    c_{k,t} \gets c_{k,t-1}+1,
\end{equation}
\begin{equation}
\label{update_mu}
     \mean{\mu_{k,t}} \gets \frac{ \mean{\mu_{k,t-1}}\cdot c_{k,t-1}+\frac{\tau_{min}}{\tau_{k,t}}}{c_{k,t}}.
\end{equation}
Then, for each client $k \in \K$, the UCB function is updated as follows:
\begin{equation}
\label{update_ucb}
    \mbox{ucb}(k,t) \gets \mean{\mu_{k,t}}+\sqrt{\frac{(m+1)\ln{t}}{c_{k,t}}}.
\end{equation}
At the initialization step, for each client $k\in\K$, $c_{k,0}$ and $\mean{\mu_{k,0}}$ are set to 0, and $\mbox{ucb}(k,0)$ is set to infinity.
Later, in the main loop, the algorithm selects clients at each iteration according to:
\vspace{-0.2cm}
\begin{equation}
\label{At}
    \al_t = \argmax_{\s \in H(\A)}\biggl\{\min_{k\in \s}{\mbox{ucb}(k,t-1)}+\frac{\alpha}{m}\sum_{k\in \s}g_{k,\Theta}\bigl(\Hi_t\bigl)\biggl\}.
\end{equation}
Afterwards, for each client $k\in \al_t$, the algorithm observes $\tau_{k,t}$ and updates $c_{k,t}$, $\mean{\mu_{k,t}}$ using (\ref{update_c}), (\ref{update_mu}).
At the end of every iteration,  $g_{k,\Theta}(\Hi_{t+1})$ and $\mbox{ucb}(k,t)$ for all client $k\in \K$ are updated using (\ref{update_ucb}).
Note that two different trade-off mechanisms are manifested during the algorithms. The first is between exploration and exploitation of the client latencies, while the second is between the iteration latency and the generalization ability.

\begin{algorithm}
\footnotesize
\caption{BSFL Algorithm}
\label{alg:Algorithm}
    \hspace*{\algorithmicindent} \textbf{Input:} Set of client indices $\K$  
    \begin{algorithmic}[1]
    \STATEx  \textbf{Initialization:}
    \STATE $\forall k\in\K: c_{k,0}\gets0, \mbox{ucb}(k,0)\gets\infty, \mean{\mu_{k,0}}\gets0$
    \STATEx  \textbf{Main loop:}
    \STATE \textbf{for} iterations $t=1, 2, ...$ \textbf{do}:
    \STATE \label{al:At} \hskip1.0em \hspace{-0.1cm}Select a set of $m$ clients using (\ref{At}) (ties are broken arbitrarily)
    \STATE \hskip1.0em Execute FL iteration
%    \STATE \hskip1.0em For each client $k\in \al_t$ if $iid=False$ and $c_{k,t}=0$ then
%    \STATEx \hskip1.0em receive:  $Q_k, |X_k|$
    \STATE \hskip1.0em For each client $k\in \al_t$ observe $\tau_{k,t}$ and update $c_{k,t}$, $\mean{\mu_{k,t}}$ using
    \STATEx \hskip1.0em (\ref{update_c}), (\ref{update_mu}).
    \STATE \hskip1.0em For each client $k\in \K$ update $g_{k,\Theta}(\Hi_{t+1})$ accordingly, and 
    \STATEx \hskip1.0em $\mbox{ucb}(k,t)$ using (\ref{update_ucb})
    \STATE \textbf{until convergence}
  \end{algorithmic}
\end{algorithm}

\subsection{Regret Analysis}
\label{ssec:regret}

In this subsection, we analyze the regret (\ref{eq:regret}) achieved by BSFL analytically, and show that it has a logarithmic order with time. Let $\Delta_{max}$ be the maximum possible difference between the expected reward obtained by genie and by any selection $\s$. 

To evaluate the regret of BSFL, we define $r(\s,\Hi_t)$ as the reward that could have been obtained at the end of iteration $t$, given the selection history $\Hi_t$, if selection $\s$ had been made. Let $\Delta_{max}$ be the maximum difference between the expected reward obtained by $\gi_t$ and by any selection $\s$, given any selection history, i.e.,
\vspace{-0.2cm}
\begin{equation}
\label{eq:delta_max}
    \Delta_{max} = \max_{\s \in H(\K), \Hi_t \subset H(\K), t\in \mathbb{N}}{ \mathbb{E}\bigr[r(\gi_t,\Hi_t)-r(\s,\Hi_t)\bigr]}.
\end{equation}
Note that $\Delta_{max}$ is bounded by: 
\begin{align*}
    \Delta_{max} & \leq 2 \alpha + \mu_{max}-\mu_{min},
\end{align*}
where $\mu_{max}$ and $\mu_{min}$ are the expected speeds of the fastest and slowest clients, respectively.
\begin{theorem}
\label{regret}
At each iteration $n$, the regret of BSFL is upper bounded by:
\vspace{-0.2cm}
\begin{equation}
    \Delta_{max} \cdot K \cdot \biggl(\frac{4(m+1)\ln{n}}{\Delta_{min}^2}+1+\frac{\pi^2}{3}\biggl).
\end{equation}
\end{theorem}
The proof can be found in Appendix \ref{ap: a}.

Theorem \ref{regret} implies that BSFL achieves a logarithmic regret order with time $O(\ln n)$. 

It is worth noting that the selection policy of Algorithm \ref{alg:Algorithm}, i.e., finding the maximum value in the update rule stated in (\ref{At}), can be computationally infeasible due to its exponential complexity. Therefore, in the following section, we will develop a practical solution to solve the maximization problem.

\section{Complexity Reduction Using A Novel Accelerated Lightweight Simulated Annealing}

The most computationally challenging aspect of the BSFL algorithm is selecting clients at the beginning of each iteration, as indicated by line 3 of the pseudocode. Specifically, the server needs to find the client selection that maximizes the expression in the update rule (\ref{At}). Classic deterministic methods for finding this selection have a time complexity of $O\binom{|\mathbb{A}_t|}{m}$, which is computationally infeasible when the number of channels ($m$) is large. Therefore, heuristic methods are often used to find good approximate solutions to optimization problems in a finite amount of time. In particular, simulated annealing (SA) is a heuristic optimization algorithm that has strong theoretical properties of convergence guarantee as the number of time steps increases. In the subsequent sections, we present a new SA-type algorithm that leverages the unique structure of the MAB-based FL scheduling problem to accelerate the stochastic optimization process through the design of a lightweight search space. The proposed SA-type method demonstrates significantly faster convergence compared to the classical method while still preserving the strong theoretical convergence guarantee property as the number of time steps used to execute line 3 of the pseudocode increases.

\subsection{The Stochastic Optimization Flow}

To implement SA-type algorithm to solve (\ref{At}), we need to define a multi-state environment, in which each state $s$ has neighboring states denoted as $N_s$. Each state also has an associated energy, represented by $E(s)$. The objective of the algorithm is to find the state with the highest energy (or, alternatively, the lowest energy, depending on the problem formulation). To facilitate movement between states, a temperature parameter $T_i$ for time step $i$ is introduced. This temperature affects the probability of transitioning from the current state to a state with lower energy. The algorithm begins by randomly selecting an initial state $s_0$. It then proceeds through a series of time steps, during which the temperature is updated and a neighboring state $u \in N_{s_i}$ is chosen at random. The next state $s_{i+1}$ is then updated according to:
\begin{equation}
s_{i+1}\gets\begin{cases}
		u, & \text{if $E(u)\geq E(s_i)$,}\\
            u, & \text{w.p., $e^{\frac{E(u)-E(s_i)}{T_i}}$ if $E(u)<E(s_i)$, }\\
            s_i, & \text{otherwise.}
		 \end{cases}
\end{equation}  
In our MAB-based FL scheduling setting, each client selection determines a state, which means that in each FL round $t$, the number of states of the stochastic optimization problem would be ${|\A_t| \choose m}$. The energy of each state would be:
\begin{equation}
    E(\s) = \min_{k\in \s}\biggr\{\mbox{ucb}(k,t)+\frac{\alpha}{m}\sum_{k\in \s}g_{k,\Theta}\bigl(\Hi_t\bigl)\biggl\}.
\end{equation}

As previously mentioned, the SA dynamics requires that each state has neighboring states. A direct implementation of the classic SA method results in a structure where any two client selections (which represent two states), $\s$ and $\mathcal{U}$, will be considered neighboring selections only if they differ in only one client. Formally,
\begin{equation}
    \s \in N_{\mathcal{U}} \;\;,\;\; \mathcal{U} \in N_{\s} \Leftrightarrow \bigl|\s \cap \mathcal{U}\bigr|=m-1,
\end{equation}
where $m$ is the number of channels, same as before.
It can be seen that through this construction, the neighborhood is symmetrical between the selections (states) and that for each selection there are $m(K-m)$ neighbors. From \cite{hajek1988cooling}, setting the temperature at each time step to $T_i = \frac{\Delta_{max}}{\log{(i+1)}}$ guarantees convergence to the selection with the maximum energy as the number of time steps in the SA algorithm approaches infinity, where $\Delta_{max}$ is defined in (\ref{eq:delta_max}). Despite the strong theoretical convergence guarantee, a recognized disadvantage of the SA method is its rate of convergence which can be quite slow. Thus, in the subsequent section, we present a novel SA-type algorithm that addresses this issue while still maintaining the theoretical convergence guarantee.   

\subsection{The Proposed Accelerated Lightweight Simulated Annealing (ALSA) algorithm}

 \begin{figure*}[t]
    \centering    \includegraphics[width=0.8\textwidth]{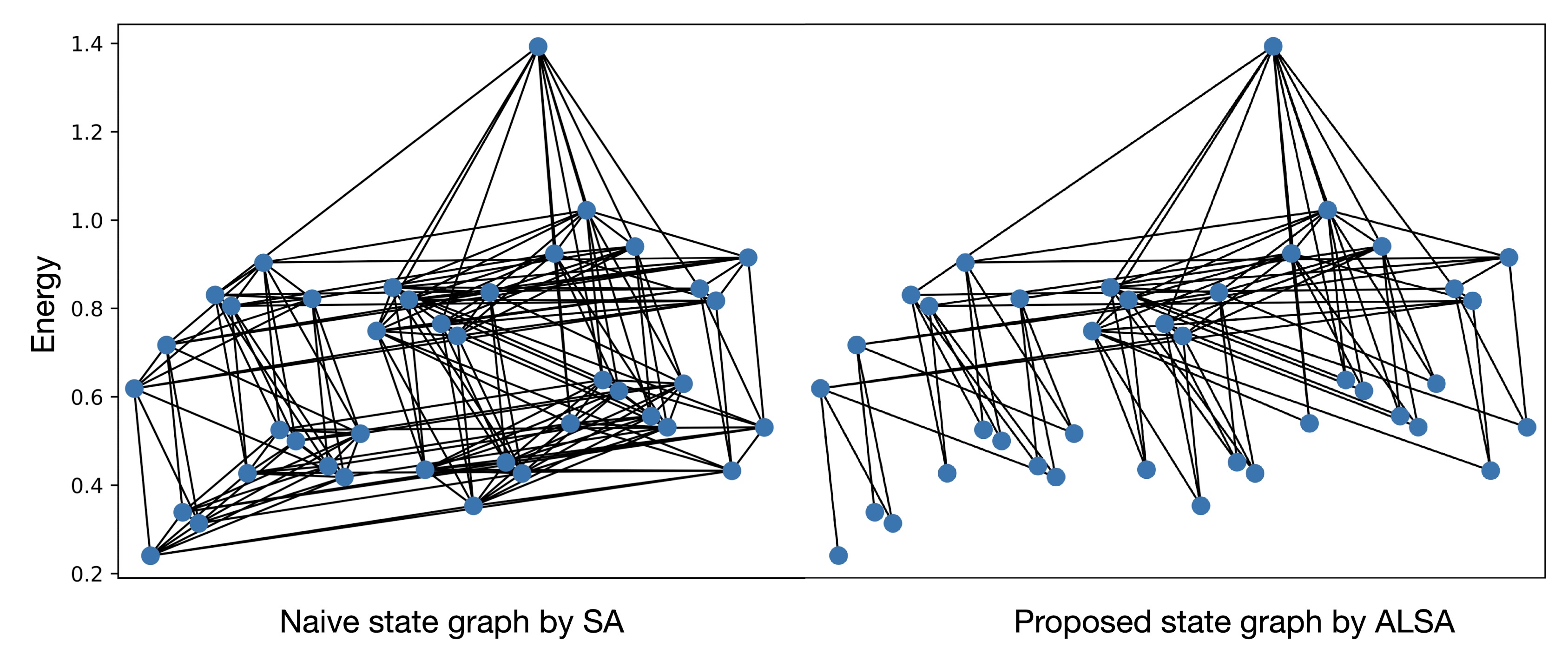}
    \vspace{-0.2cm}
    \caption{An illustration of the sub-graphs of states resulting by ALSA and SA. For clarity, only sub-graphs of the complete graphs are shown.}
    \vspace{-0.3cm}
    \label{fig:illustration}
\end{figure*}

Building on the concept of SA-type dynamics, we present a novel accelerated SA-type method that capitalizes on the specific structure of the MAB-based FL scheduling problem. Our proposed method accelerates the optimization process by designing a lightweight search space, resulting in faster convergence compared to the classic SA technique. This is achieved by taking into account the unique features of the MAB-based FL scheduling problem.

When applying the classic SA method to our problem, the number of neighbors for each state is $m(K-m)$, which becomes $O(K^2)$ when $m=O(K)$. In our proposed ALSA method, we exploit the characteristics of the multi-armed bandit optimization to reduce the number of connections between states and create a lightweight state graph for the search space. This modification allows for faster convergence while still maintaining the theoretical convergence guarantee of the SA method (which will be shown later). Specifically, we define a new neighborhood rule that only allows for client selections that differ by a single client from the current selection, and requires that this client has the lowest $\mbox{ucb}$ value or the lowest $g_{k,\Theta}$ value in one of the selections. Formally, 
\begin{multline}
    \s \in N_{\mathcal{U}} \;\;,\;\; \mathcal{U} \in N_{\s}\\    \hspace{-2cm}\Leftrightarrow \biggl(\bigl|\s \cap \mathcal{U}\bigr|=m-1\biggr) \text{  and} \\ \biggl(\s\backslash\mathcal{U}\subset \Bigl\{\argmin_{k \in \s}{\mbox{ucb}(k,t)}, \argmin_{k\in\s}{g_{k, \Theta}(\Hi_t)}\Bigr\} \text{  or} \\
    \mathcal{U}\backslash\s \subset \Bigl\{\argmin_{k \in \mathcal{U}}{\mbox{ucb}(k,t)}, \argmin_{k\in\mathcal{U}}{g_{k, \Theta}(\Hi_t)}\Bigr\}\biggr).
\end{multline}
It is important to note that each selection $\s$ has at most $2(K-m)$ other selections that differ by only one client $k$ which is in the set 
\begin{center}
$\Bigl\{\argmin_{k \in \s}{\mbox{ucb}(k,t)}, \argmin_{k\in\s}{g_{k, \Theta}(\Hi_t)}\Bigr\}$.    
\end{center} 
The neighborhoods in this formulation are symmetrical, and the average number of neighbors for each selection is no more than $4(K-m)$, resulting in a total of $O(K)$ neighbors regardless of the value of $m$.

Denote the state with the global maximum energy as $s^*$. The theoretical convergence of ALSA is shown next.

\begin{theorem}
    \label{th: SA}
    Implementing ALSA with temperature $T_i = \frac{\Delta_{max}}{\log{(i+1)}}$ at time step $i$ yields:
    \begin{equation}
       \displaystyle
       \lim_{i\to \infty}E(s_i)=E(s^*) . 
    \end{equation}
\end{theorem}
The proof can be found in Appendix \ref{ap: b}. 

Theorem \ref{th: SA} implies that as the number of time steps for running ALSA increases, the result of executing ALSA will converge to the optimal solution of (\ref{At}). By using ALSA, the execution of line 3 in the BSFL algorithm (Algorithm \ref{alg:Algorithm}) becomes efficient.

\subsection{A Comparison of SA and ALSA for solving (\ref{At})}

The classic SA method, when applied to our problem, results in a graph structure in which each state has $O(K^2)$ neighbors. However, by exploiting the characteristics of the MAB-based optimization, our proposed ALSA method is able to reduce the number of connections between states to create a lightweight graph for the search space, while still maintaining the theoretical convergence guarantee as previously demonstrated. This construction results in a graph structure in which each state has $O(K)$ neighbors.

\begin{figure}[htbp]
\centering \epsfig{file=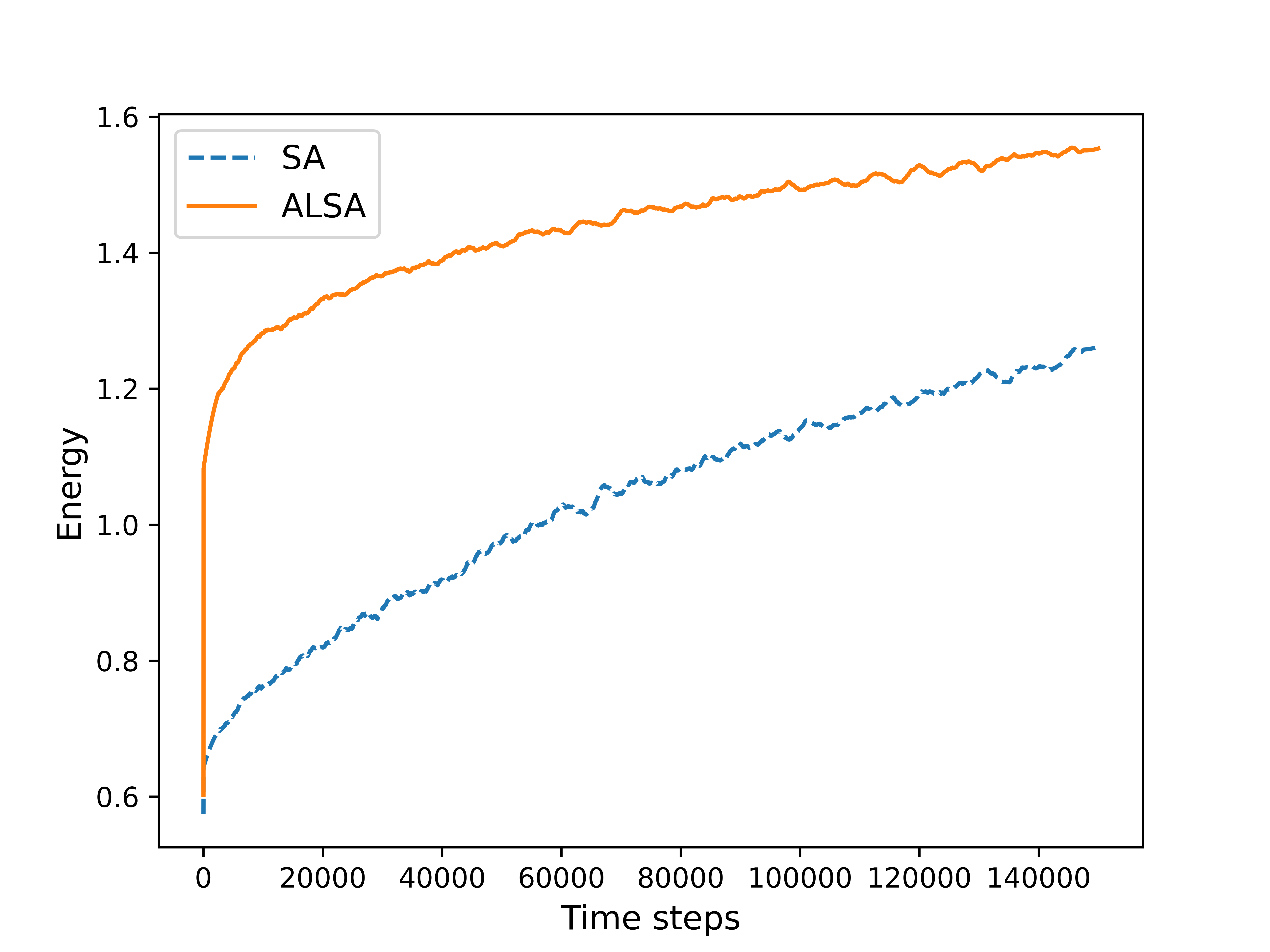,
width=0.45\textwidth}
\vspace{-0.2cm} 
\caption{The energy of the current state as a function of the time step of ALSA and SA runs is depicted in this figure. The simulation in this figure is based on the selection of $25$ clients out of $500$, for which it is computationally infeasible to directly solve 
(\ref{al:At}) ($\approx 2^{139}$ different selections).} \vspace*{-0.3cm} 
\label{fig:SA compare}
\end{figure}

To demonstrate the effectiveness of our proposed method, we conducted thousands of runs with varying numbers of clients and selection sizes. In the vast majority of runs (98.3\%), ALSA reached a state with a higher energy within the fixed time period compared to the classic SA method. Furthermore, we observed that the majority of the additional edges in the classic SA method are between low-energy states or states with similar energy, which do not contribute to the convergence to the global maximum state and instead cause the algorithm to wander for a longer time between these low-energy states. This is illustrated in Figure \ref{fig:illustration}, which shows a comparison between the graph structures created by ALSA and SA for a simulation of selecting 4 clients out of 8 with drawn $\mbox{ucb}(k,t)$ and $g_k$ values. As can be observed in Fig. \ref{fig:SA compare}, the judicious removal of excess edges in the graph created by ALSA leads to a significantly faster rate of convergence to high-energy states compared to SA.

\section{Simulation Results}

In this section, we present the results of our simulations, which include two types of datasets and models. The first is synthetic data for linear regression and the second is image data from two well-known datasets, Fashion-MNIST and CIFAR-10, for a convolutional neural network (CNN) model. 

We consider two scenarios for each simulation. In the first scenario, the data (images or synthetic data) is divided i.i.d. between the clients, and for the image databases each client has an equal number of images from each class. In the second and more challenging scenario, each client has a different amount of data and, in the image datasets, a different distribution of images among the ten classes. 

In the i.i.d. case, we compared the proposed BSFL client selection algorithm to the client selection UCB (CS-UCB) algorithm proposed in \cite{xia2020multi} and to a random selection policy \cite{mcmahan2017communication}. During each simulation, the model was trained for the same amount of time, but the different algorithms resulted in varying numbers of FL rounds (i.e., iterations) for the same amount of total training time. In other words, the number of iterations performed by each algorithm for a given latency in the figures is not the same since the running time of each iteration varies between different client-set selections.

\begin{figure}[h!]
\vspace{-0.2cm} 
\centering \epsfig{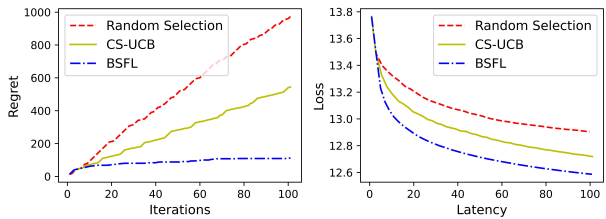}
\vspace{-0.6cm} 
\caption{Linear regression model with synthetic data in the i.i.d. scenario. Figure (left): Regret as a function of iterations. Figure (right): Test loss as a function of latency.} \vspace*{-0.1cm} 
\label{fig:lin iid}
\end{figure}

To compare the regret between the algorithms, we first had to search through all $\binom{K}{m}$ client selections in each iteration and determine which selection truly maximized (\ref{Gt}). Therefore, in order to compare the regret, we chose a relatively small number of clients (namely, 20) and a selection size of 5. As shown in Figure \ref{fig:lin iid} (left), BSFL achieves a logarithmic regret compared to the other algorithms, which reached a linear regret. Additionally, Figure \ref{fig:lin iid} (right) illustrates that BSFL achieves the smallest loss values among the algorithms.

We start by comparing the regret between the algorithms. To compare the regret, we first had to search through all $\binom{K}{m}$ client selections in each iteration and determine which selection truly maximized (\ref{Gt}), i.e., find the selection $\gi_t$. Therefore, in order to compare the regret, we chose a relatively small number of clients (namely, $20$) and a selection size of $5$. As shown in Figure \ref{fig:lin iid} (left), BSFL achieves a logarithmic regret compared to the other algorithms, which reached a linear regret. Additionally, Figure \ref{fig:lin iid} (right) illustrates that BSFL achieves the smallest loss values among the algorithms.

\begin{figure}[h!]
\vspace{-0.2cm} 
\centering \epsfig{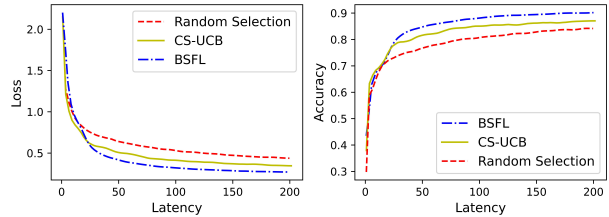}
\vspace{-0.6cm} 
\caption{CNN model with Fashion-MNIST data in the i.i.d. scenario. Figure (left): Test loss as a function of latency. Figure (right): Test accuracy as a function of latency.} \vspace*{-0.1cm} 
\label{fig:fashion_mnist iid}
\end{figure}

\begin{figure}[h!]
\vspace{-0.2cm} 
\centering \epsfig{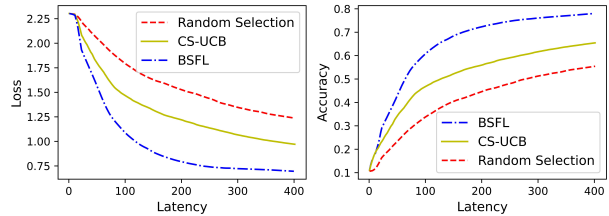}
\vspace{-0.6cm} 
\caption{CNN model with CIFAR10 data in the i.i.d. scenario. Figure (left): Test loss as a function of latency. Figure (right): Test accuracy as a function of latency.} \vspace*{-0.1cm} 
\label{fig:cifar10 iid}
\end{figure}

Second, while still operating within the i.i.d. case, we conducted simulations on real data using a CNN model with $500$ clients and a selection size of $25$. The CNN model employed a standard architecture comprising convolutional layers, max-pooling layers, dense layers, and dropout regularization. The activation function used in the layers was 'ReLU', with the exception of the final layer, where we used the softmax function. Additionally, we employed the categorical cross-entropy loss function for both models. In Figures \ref{fig:fashion_mnist iid}, \ref{fig:cifar10 iid},  we present the results of our simulations for each of the algorithms. The simulation results demonstrate the performance of the global model on test data throughout the training process. The figures show that BSFL significantly outperforms the other algorithms in terms of both loss and accuracy percentage on the test data. 

\begin{figure}[h!]
\vspace{-0.2cm} 
\centering \epsfig{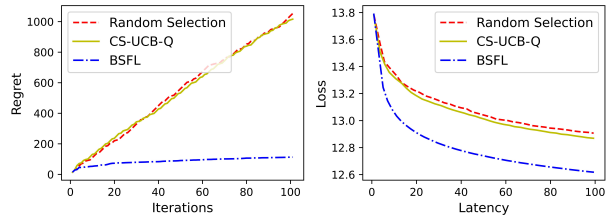}
\vspace{-0.6cm} 
\caption{Linear regression model with synthetic data in the non-i.i.d. scenario. Figure (left): Regret as a function of iterations. Figure (right): Test loss as a function of latency.} \vspace*{-0.1cm} 
\label{fig:lin non-iid}
\end{figure}

In the non-i.i.d. scenario, we compare the proposed BSFL algorithm to the CS-UCB-Q proposed in \cite{xia2020multi} for non-i.i.d. data, and to an adapted random selection policy for non-i.i.d. data that randomly selects the clients with a probability proportional to the size of each client's database \cite{li2019fair}. As before, to evaluate the regret, we divided the dataset into a small number of clients, and compared the regret of each algorithm. As shown in Figure \ref{fig:lin non-iid} (left), even in the non-i.i.d. scenario, BSFL achieves a logarithmic regret order, in contrast to the other algorithms which achieve a linear regret order. Furthermore, Figure \ref{fig:lin non-iid} (right) illustrates how these results in terms of regret lead to faster convergence in terms of the loss calculated on the test data, indicating superior generalization of the model trained using BSFL.

\begin{figure}[h!]
\vspace{-0.2cm} 
\centering \epsfig{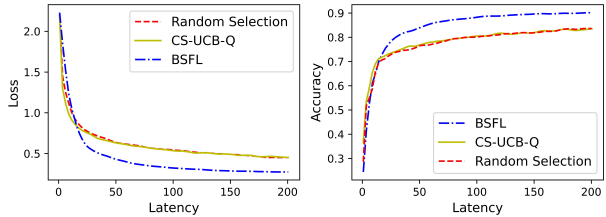}
\vspace{-0.6cm} 
\caption{CNN model with Fashion-MNIST data in the non-i.i.d. scenario. Figure (left): Test loss as a function of latency. Figure (right): Test accuracy as a function of latency.} \vspace*{-0.1cm} 
\label{fig:fashion_mnist non-iid}
\end{figure}

\begin{figure}[h!]
\vspace{-0.2cm} 
\centering \epsfig{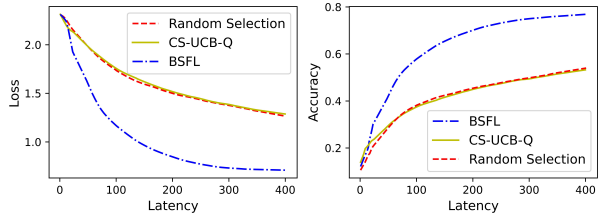}
\vspace{-0.6cm} 
\caption{CNN model with CIFAR10 data in the non-i.i.d. scenario. Figure (left): Test loss as a function of latency. Figure (right): Test accuracy as a function of latency.} \vspace*{-0.1cm} 
\label{fig:cifar10 non-iid}
\end{figure}

We conducted a similar experiment using real datasets and divided the data into a larger set of $500$ clients, with a selection size of $25$ clients per iteration. Each client contained a different number of images and varying amounts of images from each class. As shown in Figures \ref{fig:fashion_mnist non-iid} and \ref{fig:cifar10 non-iid}, in this scenario as well, BSFL leads to faster convergence in terms of both loss and accuracy on the test data. All evaluations of the global model's performance, depicted in graphs, were conducted using test data that was not utilized for training and was not accessible to any of the clients. Therefore, we can conclude that BSFL leads to superior generalization compared to the other algorithms.

These simulation results demonstrate the strong performance of BSFL for client selection in federated learning compared to existing methods.

\section{Conclusion}

We developed a novel MAB-based approach for client selection in FL systems, aimed at minimizing training latency while preserving the model's ability to generalize. We developed a novel algorithm to achieve this goal, dubbed Bandit Scheduling for FL (BSFL). BSFL was shown to achieve a logarithmic regret, defined as the difference in loss between BSFL and a genie with complete knowledge of all clients' latency means. Simulation results demonstrated that BSFL is superior to existing methods.

\appendices{}
\section{Proof of Theorem \ref{regret}}
\label{ap: a}

In this appendix we provide the proof for Theorem \ref{regret}. The regret of BSFL (\ref{eq:regret}) can be written as: 
\begin{multline}
R(n) = \mathbb{E}_\Pi\biggr[\sum_{t=1}^n{ r^*(t)-r(t)}\biggr] \\= 
\mathbb{E}\biggr[\sum_{t=1}^n{ r(\gi_t,\Hi_t)-r(\al_t,\Hi_t)}\biggr]\\=
\mathbb{E}\biggr[\sum_{t=1}^n{\Delta_{t}\cdot\mathds{1}\Bigl\{(r(\gi_t,\Hi_t)\neq r(\al_t,\Hi_t)})\Bigl\}\biggr]\\
\leq \Delta_{max} \cdot \mathbb{E}\biggr[\smashoperator[r]{\sum_{\s\in H(\mathbb{K})}}\hspace{0.2cm}{N_\s(n)}\biggr],
\end{multline}
where $\Delta_t$ is the difference between the reward achieved by the algorithm to the highest reward that could have been achieved at iteration $t$ (i.e., $\Delta_t = r(\gi_t,\Hi_t)-r(\al_t,\Hi_t) \geq 0$), and $N_\s(n)$ is the number of iterations up until the $n$th iteration in which selection $\s\in H(\mathbb{K})$ was selected and the reward given from it was strictly less than the reward that would have been received by genie's selection in the same iteration. In addition, we define a $K$-dimensional counter vector $\widetilde{\vb{N}}(n)=(\widetilde{\vb{N}}_1(n), \widetilde{\vb{N}}_2(n), ..., \widetilde{\vb{N}}_K(n))$, corresponding to the $K$ clients as follows. For each iteration, in which the selection $\al_t\in H(\mathbb{A}_t)$ achieves a lower reward than the reward of genie's selection, i.e., $r(\al_t,\Hi_t)<r(\gi_t,\Hi_t)$, then the counter of the client that has been selected the fewest number of times up to this iteration among all the selected clients that were selected in this iteration is incremented by $1$. Formally, for each client (say $i$) let $\mathcal{T}_i(n)$ denote the set of time indices up to time $n$ that satisfy the following conditions: (i) Client $i$ was selected, i.e., $i\in \al_t$ for all $t\in\mathcal{T}_i(n)$; (ii) the counter $c_{i,t}$ of client $i$ is the minimal among all selected clients, i.e., $i = \argmin_{k \in \al_t}c_{k,t}$, for all $t\in\mathcal{T}_i(n)$;
and (iii) the selection $\al_t\in H(\mathbb{A}_t)$ achieves a lower reward than genie's selection, i.e., $r(\al_t,\Hi_t)<r(\gi_t,\Hi_t)$ for all $t\in\mathcal{T}_i(n)$.  
Then, 
\begin{equation}
     \widetilde{\vb{N}}_i(n) =
     |\mathcal{T}_i(n)|.
\end{equation}

Next, we aim at upper bounding $N_\s(n)$ for each  $\s\in H(\mathbb{K})$. Note that based on the definition of $\widetilde{\vb{N}}(n)$, for every iteration in which the selection $\al_t$ has a lower reward than genie's selection, one of the coordinates in the vector $\widetilde{\vb{N}}(n)$ is incremented by 1. Therefore,
\begin{equation}
    \smashoperator[r]{\sum\limits_{\s\in H(\mathbb{K})}}\hspace{0.2cm}{N_\s(n)} = \sum_{k=1}^K \hspace{0.2cm}\widetilde{\vb{N}}_k(n),
\end{equation}
which implies
\begin{equation}
\mathbb{E}\biggr[\smashoperator[r]{\sum\limits_{\s\in H(\mathbb{K})}}\hspace{0.2cm}{N_\s(n)}\biggr] = \sum_{k=1}^K \hspace{0.2cm}\mathbb{E}\biggr[\widetilde{\vb{N}}_k(n)\biggr].
\end{equation}
Let $I_k(t)$ be the indicator for the event that $\widetilde{\vb{N}}_k(t)$ is incremented by $1$ at iteration $t$. Hence, we obtain 
\begin{multline}
\label{eq:tilde_N_i_n}
\widetilde{\vb{N}}_i(n)=\sum_{t=1}^n\mathds{1}\Bigl\{(I_i(t)=1)\Bigl\}\\
    \leq1+\sum_{t=\ceil{\frac{\K}{m}}+1}^n\mathds{1}\Bigl\{(I_i(t)=1)\Bigl\}\\
    \leq l+ \sum_{t=\ceil{\frac{\K}{m}}+1}^n\mathds{1}\Bigl\{(I_i(t)=1, \widetilde{\vb{N}}_i(t) \geq l)\Bigl\}\\
    \leq l+ \sum_{t=\ceil{\frac{\K}{m}}+1}^n\mathds{1}\Bigl\{(\min_{k\in \gi_t}{\mbox{ucb}(k,t)}+\frac{\alpha}{m}\sum_{k\in \gi_t}g_{k,\Theta}(\Hi_t)<\\
    \min_{k\in \al_t}{\mbox{ucb}(k,t)}+\frac{\alpha}{m}\sum_{k\in \al_t}g_{k,\Theta}(\Hi_t), \widetilde{\vb{N}}_i(t) \geq l)\Bigl\},
\end{multline}
where the last inequality follows since the algorithm chooses action $\al_t\neq\gi_t$ that solves (\ref{At}) although the reward is maximized by $\gi_t$. Note that according to the definition of $\widetilde{\vb{N}}$, for all $k\in\al_t$ we have: $\widetilde{\vb{N}}_i(t) \leq c_{k,t}$. Therefore, since in the indicator function there is an intersection with the event that $\widetilde{\vb{N}}_i(t) \geq l$ we have for all $k\in\al_t$ that: $l\leq\widetilde{\vb{N}}_i(t) \leq c_{k,t}$ in (\ref{eq:tilde_N_i_n}). 
Denote $h_{c_{k,t}} = \sqrt{\frac{(m+1)\ln{t}}{c_{k,t}}}$, and $\mean{\mu_{c_k}}$ denotes the sampled mean of the $\frac{\tau_{min}}{\tau_{k}}$ of client $k$ after $c_k$ observations. 
Using these notations, we can upper bound $\widetilde{\vb{N}}_i(n)$ by:
\begin{multline}
      \widetilde{\vb{N}}_i(n) \\ \leq 
        l+ \hspace{-0.1cm}\smashoperator[r]{\sum_{t=\ceil{\frac{\K}{m}}+1}^n}\hspace{0.1cm}\mathds{1}\Biggl\{\min_{k\in \gi_t}{\{ \mean{\mu_{k,t}} + h_{c_{k,t}}\}}+\frac{\alpha}{m}\sum_{k\in \gi_t}g_{k,\Theta}(\Hi_t)<\\\hspace{0.7cm}
         \min_{k\in \al_t}{\{ \mean{\mu_{k,t}} + h_{c_{k,t}}\}}+\frac{\alpha}{m}\sum_{k\in \al_t}g_{k,\Theta}(\Hi_t), \widetilde{\vb{N}}_i(t) \geq l\Biggl\} \\ 
        \hspace{-2cm}\leq l+ \sum_{t=\ceil{\frac{\K}{m}}+1}^n\mathds{1}\Biggl\{\min_{0\leq c_{\hat{k}_{1,t}},c_{\hat{k}_{2,t}},...,c_{\hat{k}_{m,t}} \leq t} \vspace{0.2cm}\\
        \biggl\{\min_{j\in \{1,...,m\}}{\{ \mean{\mu_{c_{\hat{k}_{j,t}}}} + h_{c_{\hat{k}_{j,t}}}\}}+ \frac{\alpha}{m}\sum_{j=1}^m g_{\hat{k}_{j,t},\Theta}(\Hi_t)\biggl\}< \vspace{0.2cm}\\
        \max_{l\leq c_{\Tilde{k}_{1,t}},c_{\Tilde{k}_{2,t}},...,c_{\Tilde{k}_{m,t}} \leq t}\biggl\{\min_{j\in \{1,...,m\}} {\{ \mean{\mu_{c_{\Tilde{k}_{j,t}}}} + h_{c_{\Tilde{k}_{j,t}}}\}}+\vspace{0.2cm}\\ \frac{\alpha}{m}\sum_{j=1}^m g_{\Tilde{k}_{j,t},\Theta}(\Hi_t)\biggl\}\Biggl\} \\ 
        \leq l+\sum_{t=\ceil{\frac{\K}{m}}+1}^n \sum_{c_{\hat{k}_{1,t}}=1}^t \cdot\cdot\cdot \sum_{c_{\hat{k}_{m,t}}=1}^t \sum_{c_{\Tilde{k}_{1,t}}=l}^t \cdot\cdot\cdot \sum_{c_{\Tilde{k}_{m,t}}=l}^t \cdot \\
        \mathds{1}\biggl\{\min_{j\in \{1,...,m\}}{\{ \mean{\mu_{c_{\hat{k}_{j,t}}}} + h_{c_{\hat{k}_{j,t}}}\}}+ \frac{\alpha}{m}\sum_{j=1}^m g_{\hat{k}_{j,t},\Theta}(\Hi_t)<\\
        \min_{j\in \{1,...,m\}}{\{ \mean{\mu_{c_{\Tilde{k}_{j,t}}}} + h_{c_{\Tilde{k}_{j,t}}}\}}+ \frac{\alpha}{m}\sum_{j=1}^m g_{\Tilde{k}_{j,t},\Theta}(\Hi_t)\biggl\},
\end{multline}
where $\{\hat{k}_{j,t}: 1\leq j \leq m\}$ and $\{\Tilde{k}_{j,t}: 1\leq j \leq m\}$ are the clients in genie's selection and the server's selection at iteration $t$, respectively.
Let $ \hat{k}_t'$, $ \Tilde{k}_t'$, respectively, be the clients in genie's selection and the server's selection at iteration $t$ that minimizes the expression in the upper bound we derived, i.e., 
\begin{equation}
    \hat{k}_t' = \argmin_{\hat{k}_j\in \{\hat{k}_1,...,\hat{k}_m\}}\{\mean{\mu_{c_{\hat{k}_{j,t}}}} + h_{c_{\hat{k}_{j,t}}}\},
\end{equation}
\begin{equation}
    \Tilde{k}_t' = \argmin_{\Tilde{k}_j\in \{\Tilde{k}_1,...,\Tilde{k}_m\}}\{\mean{\mu_{c_{\Tilde{k}_{j,t}}}} + h_{c_{\Tilde{k}_{j,t}}}\}.
\end{equation}
Now, we claim that the event
\begin{multline*}
\biggl\{\mean{\mu_{\hat{k}_t'}} + h_{c_{\hat{k}_t'}}+\frac{\alpha}{m}\sum_{j=1}^m g_{\hat{k}_{j,t},\Theta}(\Hi_t)< \\
\hspace*{\fill}\mean{\mu_{\Tilde{k}_t'}} + h_{c_{\Tilde{k}_t'}}+ \frac{\alpha}{m}\sum_{j=1}^m g_{\Tilde{k}_{j,t},\Theta}(\Hi_t)\biggl\}    
\end{multline*}
implies that at least one of the 3 following events must occur:
\begin{enumerate}[label=(\roman*)]
    \item \label{in A} $\mean{\mu_{\hat{k}_t'}} + h_{c_{\hat{k}_t'}}+\frac{\alpha}{m}\sum_{j=1}^m g_{\hat{k}_{j,t},\Theta}(\Hi_t)$
    
    \hfill$\leq \mu_{\hat{k}_t'}+\frac{\alpha}{m}\sum_{j=1}^m g_{\hat{k}_{j,t},\Theta}(\Hi_t)$;\vspace{0.2cm} 
    \item \label{in B}$\mean{\mu_{\Tilde{k}_t'}}+\frac{\alpha}{m}\sum_{j=1}^m g_{\Tilde{k}_{j,t},\Theta}(\Hi_t)$
    
    \hfill$\geq \mu_{\Tilde{k}_t'} + h_{c_{\Tilde{k}_t'}}+\frac{\alpha}{m}\sum_{j=1}^m g_{\Tilde{k}_{j,t},\Theta}(\Hi_t)$;\vspace{0.2cm} 
    \item \label{in C}$\mu_{\hat{k}_t'}+\frac{\alpha}{m}\sum_{j=1}^m g_{\hat{k}_{j,t},\Theta}(\Hi_t) \vspace{0.2cm}$
    
    \hfill$< \mu_{\Tilde{k}_t'} + 2h_{c_{\Tilde{k}_t'}}+\frac{\alpha}{m}\sum_{j=1}^m g_{\Tilde{k}_{j,t},\Theta}(\Hi_t)$.\vspace{0.2cm} 
\end{enumerate}
We next prove this claim by contradiction. Assume that all three inequalities do not hold. Therefore, it follows that:
\begin{multline}
    \mean{\mu_{\hat{k}_t'}} + h_{c_{\hat{k}_t'}}+\frac{\alpha}{m}\sum_{j=1}^m g_{\hat{k}_{j,t},\Theta}(\Hi_t)>\mu_{\hat{k}_t'}+\frac{\alpha}{m}\sum_{j=1}^m g_{\hat{k}_{j,t},\Theta}(\Hi_t) \\ \geq
    \mu_{\Tilde{k}_t'} + 2h_{c_{\Tilde{k}_t'}}+\frac{\alpha}{m}\sum_{j=1}^m g_{\Tilde{k}_{j,t},\Theta}(\Hi_t) \\ > \mean{\mu_{\Tilde{k}_t'}}+ h_{c_{\Tilde{k}_t'}}+\frac{\alpha}{m}\sum_{j=1}^m g_{\Tilde{k}_{j,t},\Theta}(\Hi_t),
\end{multline}
where the first transition is derived from \ref{in A}, the second from \ref{in C}, the last from \ref{in B}, and all three together contradict the event.
Now, we aim at upper bounding the probabilities $Pr$\ref{in A}, $Pr$\ref{in B} that events \ref{in A} and \ref{in B} will occur:
\begin{multline}
    Pr\mbox{\ref{in A}} = Pr\Bigl(\mean{\mu_{\hat{k}_t'}} + h_{c_{\hat{k}_t'}}+\frac{\alpha}{m}\sum_{j=1}^m g_{\hat{k}_{j,t},\Theta}(\Hi_t)\vspace{0.2cm}\\
    \leq \mu_{\hat{k}_t'}+ \frac{\alpha}{m}\sum_{j=1}^m g_{\hat{k}_{j,t},\Theta}(\Hi_t)\Bigl)\vspace{0.2cm}\\ 
    = Pr\bigl( \mean{\mu_{\hat{k}_t'}} + h_{c_{\hat{k}_t'}}\leq \mu_{\hat{k}_t'}\bigl) \vspace{0.3cm}\\
    \leq e^{-2c_{\hat{k}_t'}^2\cdot \frac{(m+1) \ln{t}}{c_{\hat{k}_t'}}\cdot\frac{1}{c_{\hat{k}_t'}}} = t^{-2(m+1)},
\end{multline}
where the inequality is due to Hoeffding's inequality.
Similarly, we can upper bound $Pr$\ref{in B} by the same upper bound and obtain:
\begin{equation}
    Pr\mbox{\ref{in B}} \leq t^{-2(m+1)}. 
\end{equation}
To ensure that \ref{in C} will not occur we need to put a lower bound on $l$ (i.e., the minimum number of times a client should be selected when he has the minimum number of selections so far among the clients in the current selection):\vspace{0.4cm}\\
$\bigl\{\mu_{\hat{k}_t'}+\frac{\alpha}{m}\sum_{j=1}^m g_{\hat{k}_{j,t},\Theta}(\Hi_t) <  \vspace{0.2cm}$\\
     \hspace*{\fill}$\mu_{\Tilde{k}_t'} +  2h_{c_{\Tilde{k}_t'}}+\frac{\alpha}{m}\sum_{j=1}^m g_{\Tilde{k}_{j,t},\Theta}(\Hi_t)\bigl\}  \Leftrightarrow  $\vspace{0.2cm}\\
    $\bigl\{\mu_{\hat{k}_t'}+\frac{\alpha}{m}\sum_{j=1}^m g_{\hat{k}_{j,t},\Theta}(\Hi_t)$\vspace{0.2cm}\\
    \hspace*{\fill}$-\mu_{\Tilde{k}_t'} -\frac{\alpha}{m}\sum_{j=1}^m g_{\Tilde{k}_{j,t},\Theta}(\Hi_t) > 2h_{c_{\Tilde{k}_t'}}\bigl\}$.\vspace{0.4cm}\\
Denote the LHS of the last inequality by $\Delta_{\gi_t,\al_t,\alpha}$. Then, for the last inequality to hold, we can demand that for every $\al_t$ and $\gi_t$ selections by the algorithm and genie, respectively (which satisfy $r(\al_t,\Hi_t)< r(\gi_t,\Hi_t)$):
\begin{equation}
    \Delta_{\gi_t,\al_t,\alpha} \geq 2\sqrt{\frac{(m+1)\ln{t}}{c_{\Tilde{k}_t'}}},
\end{equation}
and because we have already shown that $\forall k\in\al_t: l \leq c_{k,t}$ and $t\leq n$, it is sufficient to demand
\begin{equation}
    \Delta_{\gi_t,\al_t,\alpha} \geq 2\sqrt{\frac{(m+1)\ln{n}}{l}}.
\end{equation}
Therefore, for $l \geq \frac{4(m+1)\ln{n}}{\Delta_{G_t,A_t,\alpha}^2}$ for every $t$, or alternatively, we can choose $l = \ceil[\Big]{\frac{4(m+1)\ln{n}}{\Delta_{min}^2}}$\vspace{0.1cm} and obtain that inequality \ref{in C} will not be met. Hence, only one of the first two inequalities must occur, and we obtain
\begin{equation}
\begin{array}{l}
\displaystyle
\mathbb{E}\biggl[\widetilde{\vb{N}}_i(n)\biggl] \vspace{0.2cm}\\
\displaystyle
\leq 
l+\sum\limits_{t=\ceil{\frac{\K}{m}}+1}^n \sum\limits_{c_{\hat{k}_{1,t}}=1}^t \cdot\cdot\cdot \sum\limits_{c_{\hat{k}_{m,t}}=1}^t \sum\limits_{c_{\Tilde{k}_{1,t}}=l}^t \cdot\cdot \cdot \sum\limits_{c_{\Tilde{k}_{m,t}}=l}^t 
\vspace{0.2cm}\\\displaystyle\hspace{5.5cm}
\left(Pr\ref{in A}+Pr\ref{in B}\right)
\vspace{0.2cm}\\
\displaystyle
 \leq  \ceil[\Big]{\frac{4(m+1)\ln{n}}{\Delta_{min}^2}}\vspace{0.2cm}\\
\displaystyle
+\sum\limits_{t=\ceil{\frac{\K}{m}}+1}^\infty 
\sum\limits_{c_{\hat{k}_{1,t}}=1}^t \cdot\cdot\cdot \sum\limits_{c_{\hat{k}_{m,t}}=1}^t \sum\limits_{c_{\Tilde{k}_{1,t}}=1}^t \cdot\cdot\cdot \sum\limits_{c_{\Tilde{k}_{m,t}}=1}^t 
\vspace{0.2cm}\\\displaystyle\hspace{6.5cm}
2t^{-2(m+1)} 
\vspace{0.2cm}\\
\displaystyle
\leq  \frac{4(m+1)\ln{t}}{\Delta_{min}^2}+1+\frac{\pi^2}{3}.
\end{array}    
\end{equation}
Finally, we can upper bound the regret by:
$$R(n) \leq \Delta_{max} K \cdot \biggl(\frac{4(m+1)\ln{n}}{\Delta_{min}^2}+1+\frac{\pi^2}{3}\biggl).$$
\hspace{7.9cm}$\square$

\section{Proof of Theorem \ref{th: SA}}
\label{ap: b}
In this appendix we provide the proof for Theorem \ref{th: SA}.
From \cite{hajek1988cooling}, the following conditions are sufficient to guarantee the convergence of cooling procedure to the state with the lowest energy
(or, alternatively, the highest energy, depending on the problem formulation):
\begin{enumerate}[label=(\roman*)]
    \item \label{it:WR} Weak Reversibility: For any energy $E$ and any two states $s_1$ and $s_2$, $s_1$ is reachable at height energy $E$ from state $s_2$, i.e., there exists a path from $s_1$ to $s_2$ that goes only through states with energy $E$ or higher) iff $s_2$ is reachable from $s_1$ at height $E$.
    \item \label{it:cooling} The temperature is set to $T_i=\frac{d}{\log{(i+1)}}$, where $d$ is greater then the difference between the energies of the highest local maxima and the minimum energy state.
    \item \label{it:irred} The Markov chain is irreducible.
\end{enumerate}
Next, we prove that all conditions are met by ALSA. Condition \ref{it:WR} follows immediately by the definition of the neighborhoods which is symmetrical, i.e., in a graph with symmetric neighborhoods, any path from one node to another can also be in the opposite direction and go through the exact same states. Regarding condition \ref{it:cooling}, since $\Delta_{max}$ is defined as the largest possible energy difference between any two states, it is in particular greater than the difference in energies between any local maximum and the state with the minimum energy. Therefore, condition \ref{it:cooling} also holds.

Finally, we need to show that condition \ref{it:irred} holds. We will show this by proving that in the newly structured state graph by ALSA, from every possible state there exists a path that reaches $s^*$, and due to the symmetric neighborhoods, this will complete the proof. Denote $s_0$ as some arbitrary state (selection). Define the state $s_1$ to be $s_1 = \bigl(s_0\backslash\{\argmin_{k \in \s_0}{\mbox{ucb}(k,t)}\}\bigr)\cup\{\argmin_{k \in \s^*}{\mbox{ucb}(k,t)}\}$. Note that $s_1$ and $s_0$ are neighboring states. For $j=1,2,3,...$ let us define the rest of the path with two phases (P1,P2) as follows:
\begin{itemize}
    \item[(P1)] As long as\vspace{0.2cm}\\
    $\argmin_{k \in \s^*}{\mbox{ucb}(k,t)} \neq \argmin_{k \in \s_j}{\mbox{ucb}(k,t)}$:\vspace{0.2cm}
    
    \hspace{0.5cm} define $s_{j+1} =   \bigl(s_j\backslash\{\argmin_{k \in \s_j}{\mbox{ucb}(k,t)}\}\bigr)
    \vspace{0.2cm}$
    
    \hfill $\cup\bigl\{\argmax_{k \in \s^*}{\mbox{ucb}(k,t)}\bigl\}$. \vspace{0.2cm}
    \item[(P2)] After Phase 1 ends, as long as $s^*\neq s_j$:\vspace{0.2cm}
    
     \hspace{0.5cm} define $s_{j+1} = \bigl(s_j\backslash\{\argmin_{k \in \s_0}{g_{k,\Theta}\bigl(\Hi_t\bigl)}\}\bigr)\vspace{0.2cm}$
    
    \hfill $\cup\{\argmax_{k \in \s^*}{g_{k,\Theta}\bigl(\Hi_t\bigl)}\}.$\vspace{0.2cm}
\end{itemize}
Note that each one of the phases lasts a finite amount of iterations, i.e., $\exists n\in \mathbb{N}: s_n=s^*$.
Note that for every $j\in\mathbb{N}$ the state $s_j$ and $s_{j+1}$ are neighbors, which implies that $P=(s_0,s_1,...,s_n=s^*)$ defined by the last phases is a path from $s_0$ to $s^*$, which completes the proof.$\hspace{1cm}\square$

\bibliographystyle{ieeetr}
\bibliography{bibliography}

\end{document}